%% file: root.tex
\documentclass[letterpaper, 10 pt, conference]{ieeeconf}     % Use this line for a4 paper

\IEEEoverridecommandlockouts                              % This command is only needed if 
\usepackage{xcolor}
\usepackage{soul}
\usepackage[utf8]{inputenc}
\usepackage[small]{caption}
\usepackage{booktabs}
\usepackage{graphics} % for pdf, bitmapped graphics files
\usepackage{tabularx}
\usepackage{amsmath} % assumes amsmath package installed
\usepackage{amssymb}  % assumes amsmath package installed
\usepackage{mathrsfs}
\usepackage{subcaption}
\usepackage{tikz}
\usetikzlibrary{shapes,arrows, automata}
\usepackage{pgfplots}
\usepackage[per-mode=symbol]{siunitx}
\DeclareSIUnit{\mph}{mph}
\usepackage{url}
\usepackage[noadjust]{cite}
\usepackage{numprint}
\usepackage{bm}

\pgfplotsset{compat=newest}
\pgfplotsset{every axis/.append style={
	font=\LARGE}
}

\DeclareUnicodeCharacter{2212}{−}
\usepgfplotslibrary{groupplots,dateplot}
\usetikzlibrary{patterns,shapes.arrows}
\pgfplotsset{every axis legend/.append style={legend cell align=left}}
\def\man#1;{%
    \begin{scope}[shift={#1}]
        \fill [rounded corners=1.5] (0,0.4) -- (0,0.8) -- (0.4,0.8) -- (0.4,0.4) --
            (0.325,0.4) -- (0.325,0.7) -- (0.3,0.7) -- (0.3,0) -- (0.225,0) --
            (0.225,0.4) -- (0.175,0.4) -- (0.175,0) -- (0.1,0) -- (0.1,0.7) --
            (0.075,0.7) -- (0.075,0.4) -- cycle;
        \fill (0.2,0.9) circle (0.1);
    \end{scope}}

\usetikzlibrary{arrows.meta, calc, shapes}
\tikzset{%
  >={Latex[width=2mm,length=2mm]},
            base/.style = {rectangle, rounded corners, draw=black,
                           minimum width=1cm, minimum height=1cm,
                           text centered, font=\sffamily},
            simulator/.style = {base, fill=green!30, minimum width=4cm},
            solver/.style = {base, fill=red!30},
            reward/.style = {base, minimum height=1.5cm},
            module/.style = {base, minimum width=2.5cm, minimum height=1.5cm, fill=blue!30},
            module2/.style = {base, minimum width=2.5cm, minimum height=1.5cm, fill=white},
            network/.style = {base, minimum width=2.5cm, minimum height=1.5cm, fill=white},
            state/.style = {base, minimum width=0.5cm, minimum height=1.0cm, fill=white},
            io/.style = {base, minimum width=0.5cm, minimum height=1.0cm, fill=white},
}
\pgfdeclarelayer{background}
\pgfdeclarelayer{foreground}
\pgfsetlayers{background,main,foreground}

\newif\ifcuboidshade
\newif\ifcuboidemphedge

\tikzset{
  cuboid/.is family,
  cuboid,
  shiftx/.initial=0,
  shifty/.initial=0,
  dimx/.initial=3,
  dimy/.initial=3,
  dimz/.initial=3,
  scale/.initial=1,
  densityx/.initial=1,
  densityy/.initial=1,
  densityz/.initial=1,
  rotation/.initial=0,
  anglex/.initial=0,
  angley/.initial=90,
  anglez/.initial=225,
  scalex/.initial=1,
  scaley/.initial=1,
  scalez/.initial=0.5,
  front/.style={draw=black,fill=white},
  top/.style={draw=black,fill=white},
  right/.style={draw=black,fill=white},
  shade/.is if=cuboidshade,
  shadecolordark/.initial=black,
  shadecolorlight/.initial=white,
  shadeopacity/.initial=0.15,
  shadesamples/.initial=16,
  emphedge/.is if=cuboidemphedge,
  emphstyle/.style={thick},
}

\makeatother

\makeatletter
\let\NAT@parse\undefined
\makeatother
\usepackage{hyperref} 
\usepackage[capitalize]{cleveref}
\crefformat{equation}{(#2#1#3)}

\usepackage[keeplastbox]{flushend}

\title{\Large \bf Finding Failures in High-Fidelity Simulation using \\Adaptive Stress Testing and the Backward Algorithm}
\author{Mark Koren$^{1}$, Ahmed Nassar$^{2}$, and Mykel J. Kochenderfer$^{1}$% <-this % stops a space
\thanks{$^{1}$Mark Koren and Mykel J. Kochenderfer are with Aeronautics and Astronautics, Stanford University, Stanford, CA 94305, USA
        {\tt\small \{mkoren, mykel\}@stanford.edu}}
\thanks{$^{2}$Ahmed Nassar is with NVIDIA, Santa Clara, CA 95051, USA
        {\tt\small anassar@nvidia.com}}
}

\begin{document}

\maketitle
\thispagestyle{empty}
\pagestyle{empty}

%%%%%%%%%%%%%%%%%%%%%%%%%%%%%%%%%%%%%%%%%%%%%%%%%%%%%%%%%%%%%%%%%%%%%%%%%%%%%%%%
\begin{abstract}  % put your abstract here!
Validating the safety of autonomous systems generally requires the use of high-fidelity simulators that adequately capture the variability of real-world scenarios. 
However, it is generally not feasible to exhaustively search the space of simulation scenarios for failures. 
Adaptive stress testing (AST) is a method that uses reinforcement learning to find the most likely failure of a system.
AST with a deep reinforcement learning solver has been shown to be effective in finding failures across a range of different systems.
This approach generally involves running many simulations, which can be very expensive when using a high-fidelity simulator.
To improve efficiency, we present a method that first finds failures in a low-fidelity simulator.
It then uses the backward algorithm, which trains a deep neural network policy using a single expert demonstration, to adapt the low-fidelity failures to high-fidelity.
We have created a series of autonomous vehicle validation case studies that represent some of the ways low-fidelity and high-fidelity simulators can differ, such as time discretization.
We demonstrate in a variety of case studies that this new AST approach is able to find failures with significantly fewer high-fidelity simulation steps than are needed when just running AST directly in high-fidelity.
As a proof of concept, we also demonstrate AST on NVIDIA's DriveSim simulator, an industry state-of-the-art high-fidelity simulator for finding failures in autonomous vehicles.
\end{abstract}

%%%%%%%%%%%%%%%%%%%%%%%%%%%%%%%%%%%%%%%%%%%%%%%%%%%%%%%%%%%%%%%%%%%%%%%%%%%%%%%%
\section{Introduction}\label{sec:intro}
Validating the safety of autonomous systems generally requires analysis using high-fidelity (hifi) simulators~\cite{Koopman2018, Kalra2016}.
Hifi simulators generally involve complex models of the sensors and dynamics as well as execution of the full autonomy stack.
In contrast, a low-fidelity (lofi) simulator would be one that lacks most or all of these features.
Proper validation requires adequately capturing the full variability of real-world scenarios, which makes an exhaustive search for failures infeasible.

Recent work has studied the use of reinforcement learning (RL) to search the space of possible simulations to find a failure~\cite{zhang2018two, akazaki2018falsification,corso2020survey}. One such approach is known as adaptive stress testing (AST)~\cite{lee2020adaptive}.
In AST, the problem of finding the most likely failure is formulated as a Markov decision process, allowing standard RL methods to be used to efficiently find realistic failures~\cite{koren2019adaptive}. 
Over time, the RL agent learns both how to adversarially force failures in the system and how to make those failures more realistic with respect to a likelihood model. 
However, RL still requires many iterations, which can make running AST in hifi intractable.

To mitigate the computational challenges of running AST with a hifi simulator, we propose using the backward algorithm (BA)~\cite{salimans2018learning}, an algorithm for hard exploration problems that trains a deep neural network policy based on a single expert demonstration. 
The idea is to first run AST in low-fidelity (lofi) to quickly find candidate failures.
Candidate failures may be the result of faults in the system under test (SUT), but they may also be spurious failures due to the inaccuracy of lofi simulation.
We use the lofi candidate failures as the expert demonstration for the backward algorithm. 

The backward algorithm learns from an expert demonstration, but also allows stochasticity during training such that the policy can actually improve upon the expert.
As a result, we will be able to overcome fidelity-dependent differences in failure trajectories where the candidate failure does not result in a failure in hifi but is still similar to a corresponding hifi failure. 
In addition, using the lofi failures as expert demonstrations can significantly reduce the amount of hifi simulations. %, while the properties of the backward algorithm allow us to efficiently reject spurious errors. 

In summary, this paper presents a new approach with the following features:
\begin{enumerate}
    \item It uses data from low-fidelity iterations to reduce the number of training iterations that need to be run in high-fidelity.
    \item It learns to overcome fidelity-dependent differences between the low-fidelity failure trajectory and its high-fidelity counterpart.
    \item It rejects spurious errors with minimal computational cost.
\end{enumerate}
These contributions increase the computational feasibility of using AST when validating autonomous systems in high-fidelity simulation.

\begin{figure}[t]
	\centering
    % \vspace*{0.25cm}
    % \centering
    \includegraphics[width=0.95\columnwidth]{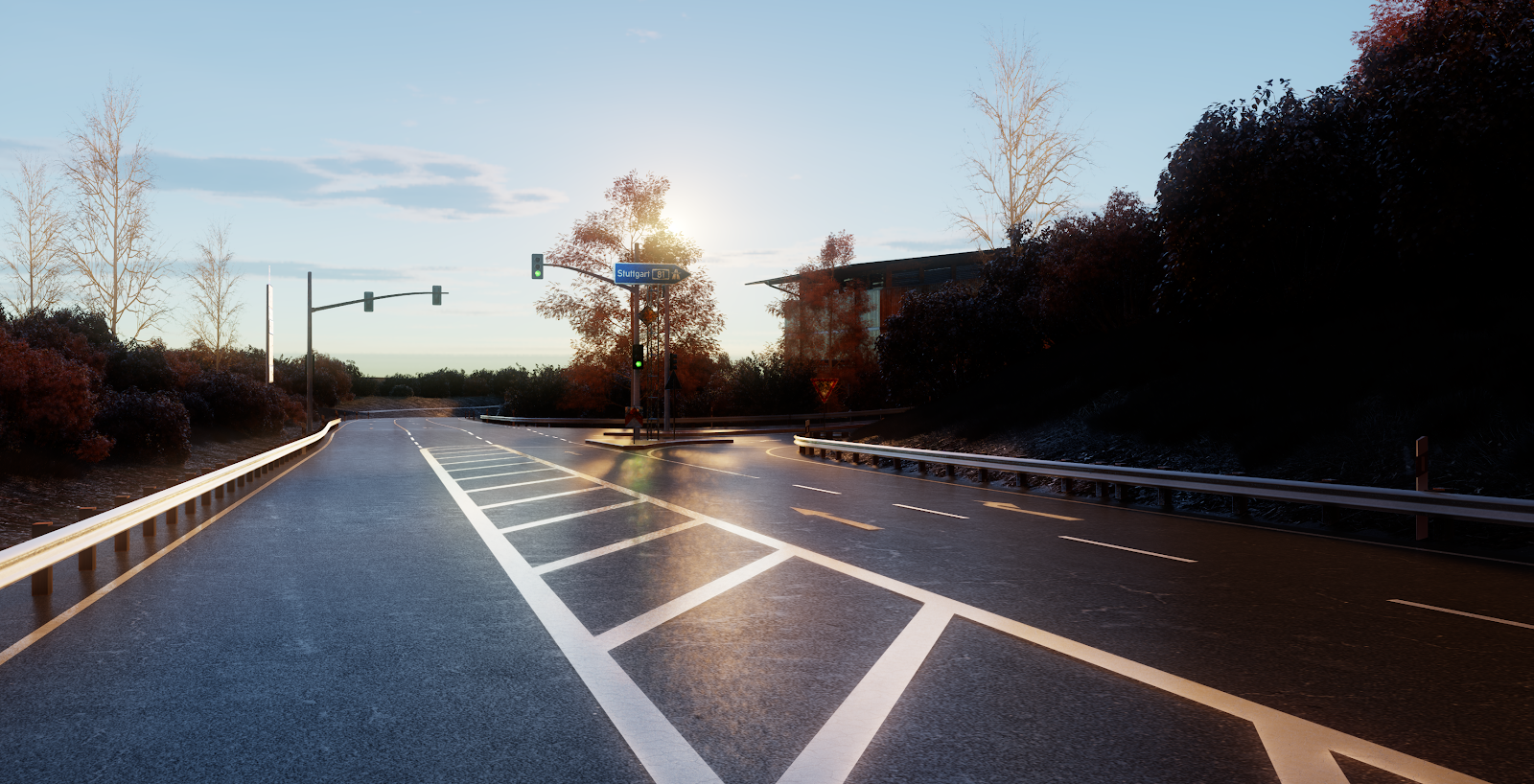}
    \caption{Example rendering of an intersection from NVIDIA's Drivesim simulator, an industry example of a high-fidelity simulator.}
	\label{fig:drivesim} 
\end{figure}

This work is organized as follows. 
\Cref{sec:background} outlines the underlying concepts of RL and AST.
\Cref{sec:methodology} explains the use of the backward algorithm, as well as the modifications needed to make the BA applicable.
\Cref{sec:case_studies} provides a series of case studies that demonstrate the performance of our approach.

\section{Adaptive Stress Testing}\label{sec:background}

\subsection{Markov Decision Process}

Adaptive stress testing (AST) frames the problem of finding the most likely failure as a Markov decision process (MDP)~\cite{DMU}. 
In an MDP, an agent takes action $a$ while in state $s$ at each timestep. 
The agent may receive a reward from the environment according to the reward function $R(s,a)$. 
The agent then transitions to the next state $s'$ according to the transition probability $P(s_{t+1} \mid a_t, s_t)$. 
Both the reward and transition functions may be deterministic or stochastic. The Markov assumption requires that the next state and reward be independent of the past history conditioned on the current state-action pair $(s,a)$. 
An agent's behavior is specified by a policy $\pi(s)$ that maps states to actions, either stochastically or deterministically. 
An optimal policy is one that maximizes expected reward.
Reinforcement learning is one way to approximately optimize policies in large MDPs. 

\subsection{Formulation}

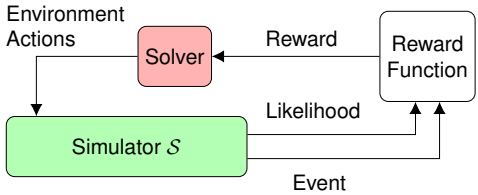
\begin{figure}[t]
	\centering
    \scalebox{0.8}{\input{Images/ASTStruct.tex}}
    \caption{The AST methodology. The simulator is treated as a black box. The solver optimizes a reward based on transition likelihood and on whether an event has occurred.}
	\label{fig:ASTStruct} 
\end{figure}

We are trying to find the most likely failure of a system, which can be seen as an optimization problem in which we are trying to maximize the probability of a trajectory such that the trajectory ends in a target subset of the state space $E$, 
\begin{equation}
\label{eq:ast_optimzation}
\begin{aligned}
& \underset{a_0, \ldots, a_t, t}{\text{maximize}}
& & P(s_0, a_0, \ldots,s_t, a_t) \\
& \text{subject to}
& & s_t \in E,
\end{aligned}
\end{equation}
where $P(s_0, a_0, \ldots,s_t, a_t)$ is the probability of a trajectory in simulator $\mathcal{S}$. Because of the Markov assumption, $s_t$ is only a function of $a_t$ and $s_{t-1}$. 
The set $E$ defines which parts of the state space constitute a failure\textemdash in the autonomous vehicle case, for example, failures could be collisions or near-misses. 

When using reinforcement learning, we maximize
\begin{equation}
\label{eq:sum_reward}
\mathbb{E}\left[\sum_{t=0}^{T} R(s_t, a_t)\right],
\end{equation}
where, according to the AST reward formulation, 
\begin{equation}
 \label{eq:base_reward}
R\left(s_t, a_t\right) = \left\{
        \begin{array}{ll}
            0 & \text{if }  s_t \in E \\
            -\infty &  \text{if } s_t \notin E, t\geq T \\
            \log P(a_t \mid s_t)   &  \text{if } s_t \notin E, t < T, % P(s_t \mid a_{t-1}, s_{t-1}) 
        \end{array}
    \right.
\end{equation}
and $T$ is the horizon~\cite{koren2019adaptive}. 
We are summing log-probabilities at each timestep, which under maximization is equivalent to multiplying probabilities at each timestep.
As a result, the trajectory that maximizes \cref{eq:sum_reward} is the same trajectory that maximizes \cref{eq:ast_optimzation}. 

The AST process is shown in \cref{fig:ASTStruct}. The solver takes environment actions, which deterministically control the simulator timestep. 
The simulator, which is treated as a black box, outputs an indicator of a failure event, as well as the likelihood of the timestep.
Both of these values are passed to the reward function to calculate a reward, which is then used by the solver to optimize the adversarial AST agent and improve its performance. 

% The solver treats both the system under test and the simulator itself as a black box, but the following access functions must be provided by the simulator:
% \begin{itemize}
%     \item \textsc{Initialize}$(\mathcal{S}, s_0)$: Resets $\mathcal{S}$ to a given initial state $s_0$.
%     \item \textsc{Step}$(\mathcal{S}, E, a)$: Steps the simulation in time by drawing the next state, $s'$, after taking action $a$. The function returns the probability of the action taken and an indicator of whether or not $s'$ is in $E$.
%     \item \textsc{IsTerminal}$(\mathcal{S}, E)$: Returns true if the current state of the simulation is in $E$ or if the horizon of the simulation $T$ has been reached.
% \end{itemize}

Previous papers have presented solvers based on Monte Carlo tree search~\cite{Koren}, deep reinforcement learning (DRL)~\cite{koren2019efficient}, and go-explore~\cite{koren2020adaptive}.
In this paper, we will use DRL and the BA, with background provided for those unfamiliar with either approach.

\subsection{Deep Reinforcement Learning}

In deep reinforcement learning (DRL), a policy is represented by a neural network~\cite{goodfellow2016deep}.
Whereas a feed-forward neural network maps an input to an output, we use a recurrent neural network (RNN), which maps an input and a hidden state from the previous timestep to an output and an updated hidden state. 
An RNN is naturally suited to sequential data due to the hidden state, which is a learned latent representation of the current state.
RNNs suffer from exploding or vanishing gradients, a problem addressed by variations such as long-short term memory (LSTM)~\cite{Hochreiter1997} or gated recurrent unit (GRU)~\cite{bahdanau2014neural} networks.

There are many different algorithms for optimizing a neural network, proximal policy optimization (PPO)~\cite{schulman2017proximal} being one of the most popular. 
PPO is a policy-gradient method that updates the network parameters to minimize the cost function.
Improvement in a policy, compared to the old policy, is measured by an advantage function, which can be estimated for a batch of rollout trajectories using methods such as generalized advantage estimation (GAE)~\cite{Schulman2015}.
However, variance in the advantage estimate can lead to poor performance if the policy changes too much in a single step. 
To prevent such a step leading to a collapse in training, PPO can limit the step size in two ways: 1) by incorporating a penalty proportional to the KL-divergence between the new and old policy or 2) by clipping the estimated advantage function when the new and old policies are too different.

\subsection{Backward Algorithm}

The backward algorithm is an algorithm for hard exploration problems that trains a deep neural network policy based on a single expert demonstration~\cite{salimans2018learning}. 
Given a trajectory ${(s_t,a_t,r_t, s_{t+1})}_{t=0}^T$ as the expert demonstration, training of the policy begins with episodes starting from $s_{\tau_1}$, where $\tau_1$ is near the end of the trajectory. 
Training proceeds until the agent receives as much or more reward than the expert from $s_{\tau_1}$, after which the starting point moves back along the expert demonstration to $s_{\tau_2}$, where $0 \leq \tau_2 \leq \tau_1 \leq T$.
Training continues in this way until $s_{\tau_N} = s_0$. 
Diversity can be introduced by starting episodes from a small set of timesteps around $s_{\tau}$, or by adding small random perturbations to $s_{\tau}$ when instantiating an episode. 
Training based on the episode can be done with any relevant deep reinforcement learning algorithm that allows optimization from batches of trajectories, such as PPO with GAE. 

\section{Methodology}\label{sec:methodology}

While AST has been shown to find failures in autonomous systems acting in high-dimensional environments, thousands of iterations may still be required due to AST’s use of reinforcement learning. 
In order to use AST with a hifi simulator, we must find a way to reduce the number of iterations that must be run at high fidelity. 
Consequently, we first run AST in low fidelity, where we are less constrained in the number of iterations we can run. 
We then find failures in hifi using the backward algorithm with candidate failures as expert demonstrations. 
The backward algorithm has several advantages that make it well suited for this task.

The backward algorithm is able to produce a policy that improves upon the original expert demonstration. 
Even though each training episode starts from somewhere along the expert demonstration, stochasticity during training allows the policy to deviate from the expert demonstration, and the policy can learn differing behavior if such behavior leads to higher rewards.
In our case, some candidate failures may be very similar but not exactly identical to a corresponding failure in hifi. 
The backward algorithm provides structure to help search the simulation-space close to the corresponding hifi failure but allows enough stochasticity during training to still identify the hifi failure. 
This approach could potentially work even when the lofi and hifi simulators have slightly differing state spaces, for example if the lofi simulator’s perception system is a much simpler abstraction of the hifi simulator’s perception system (see \cref{sec:cs_tracker,sec:cs_perception}). 
In this case, the expert demonstration would not include the perception actions, but the agent could still learn to manipulate the perception system to force failures, albeit at an increased computational cost.

A concern when validating in lofi is the occurrence of spurious failures, which are failures that are not actually present in our autonomous system but only occur due to the poor accuracy of lofi simulator.
Another advantage of the backward algorithm is that spurious failures will be computationally cheap to reject. 
% Another advantage of the backward algorithm is that spurious errors, which are failures that are not actually present in our autonomous system but only occur due to the poor accuracy of lofi simulator, will be computationally cheap to reject. 
By putting a limit on the number of epochs per step of the expert demonstration, we can efficiently identify when the algorithm is unable to produce a failure similar to that of the expert demonstration and move on to a new candidate failure. 
Because the backward algorithm starts by training policies from the end of the expert demonstration, initial epochs have short rollout horizons. 
While some spurious errors may not be identified until late in the expert demonstration, enough will be identified early on to save a significant amount of compute time. 

Its ability to save compute time is the final advantage of the backward algorithm. 
By using lessons learned in lofi, we are able to transform a single hard problem into a series of much easier problems. 
Each step of training in the backward algorithm can be thought of as training a one-step policy, since we have already learned a policy for the rest of the trajectory. 
Although this conceptualization breaks down as problems increase in difficulty\textemdash since we may need to deviate from the expert demonstration quite a bit\textemdash the backward algorithm nevertheless can save a significant number of iterations, and many of the iterations require fewer hifi steps as well. 

Applying this approach required some tweaks to the backward algorithm. 
The original algorithm calls for training at each timestep of the expert demonstration until the policy is as good or better than the original. 
First, we relax this rule and, instead, move $s_{\tau}$ back along the expert demonstration any time a failure is found within the current epoch.
Second, we add an additional constraint of a maximum number of epochs at each timestep. 
If the policy reaches the maximum number of epochs without finding failures, training continues from the next step of the expert demonstration. 
However, if $s_{\tau}$ is moved back without finding a failure five consecutive times, the expert demonstration is rejected as a spurious error. 
We found that there are times where the BA is unable to find failures in hifi in the early stages of training due to the lofi trajectory being too dissimilar to any failure in hifi, so this change allows the BA more epochs to train and find failures.
In a similar vein, we also start training with $\tau > 0$ and when moving back along the expert demonstration after a failure we increment $\tau$ more than one step back.

\section{Case Studies }\label{sec:case_studies}

To demonstrate the BA's ability to adapt lofi failures to hifi failures, we constructed a series of case studies that represent a variety of differences one might see between lofi and hifi simulators. 
These case studies measure which types of fidelity differences the BA can handle well, and which types the BA will struggle with. 
Because the BA starts many epochs from points along the expert demonstration, many rollouts will have shorter trajectory lengths.
Therefore, a direct comparison to DRL of iterations would not be fair.
Instead, we measure performance in terms of the number of simulation steps, assuming this would be the bottleneck in hifi simulators.
Unless otherwise noted, all case studies share the following setup.

\begin{figure}[t]
	\centering
    \vspace*{0.25cm}
    % \centering
    \includegraphics[width=0.80\columnwidth]{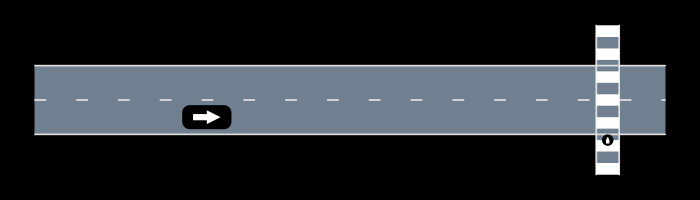}
    \caption{Layout of a running example. A car approaches a crosswalk on a neighborhood road with one lane in each direction. A pedestrian is attempting to cross the street at the crosswalk.}
	\label{fig:scenario1} 
\end{figure}

\subsubsection{Simulation}

In the test scenario, the system under test (SUT) is approaching a crosswalk on a neighborhood road where a pedestrian is trying to cross, as shown in \cref{fig:scenario1}.
The pedestrian starts $\SI{1.85}{\meter}$ back from the center of the SUT's lane, exactly at the edge of the street, and is moving across the crosswalk with an initial velocity of $\SI{1.0}{\meter\per\second}$. 
The SUT starts at $\SI{55}{\meter}$ away from the crosswalk with an initial velocity of $\SI{11.17}{\meter\per\second}$ ($\SI{25}{\mph}$), which is also the desired velocity. 
The SUT is a modified version of the intelligent driver model (IDM)~\cite{PhysRevE621805}. 
IDM is a lane following model that calculates acceleration based on factors including the desired velocity, the headway to the vehicle in front, and the IDM's velocity relative to the vehicle in front.  
Our modified IDM ignores pedestrians that are not in the street, but treats the pedestrian as a vehicle when it is in the street, which\textemdash due to large differences in relative velocity\textemdash will cause the IDM to brake aggressively to avoid collision. 
Simulation was performed with the AST Toolbox.\footnote{ \url{github.com/sisl/AdaptiveStressTestingToolbox}} 

\subsubsection{Algorithms}

To find collisions, AST was first run with a DRL solver in each case study's low fidelity version of the simulator.  
Once a collision was found, the backward algorithm was run using the lofi failure as the expert demonstration.  
Results are shown both for instantiating the backward algorithm's policy from scratch and for loading the policy trained in lofi. 
Results are compared against running AST with the DRL solver from scratch in hifi. 
Optimization for all methods is done with PPO and GAE, using a batch size of \SI{5000}{}, a learning rate of \SI{1.0}{}, a maximum KL divergence of \SI{1.0}{}, and a discount factor of \SI{0.99}{}. 
The BA starts training \SI{10}{} steps back from the last step, and moves back \SI{4}{} steps every time a failure is found during a batch of rollouts.

\subsection{Case Study: Time Discretization} \label{sec:cs_time}

In this case study, the fidelity difference is time discretization and trajectory length. 
The lofi simulator runs with a timestep of \SI{0.5}{} seconds for \SI{10}{} steps, while the hifi simulator runs with a timestep of \SI{0.1}{} seconds for \SI{50}{} steps.
This fidelity difference approximates skipping frames or steps to reduce runtime.
In order to get an expert demonstration of the correct length and discretization in hifi, the lofi actions were repeated \SI{5}{} times for each lofi step. 
The results are shown in \cref{table:time_results}.

The hifi DRL baseline took \SI{44800}{} simulation steps to find a failure.
The lofi solver was run for \SI{5}{} epochs, finding a failure after \SI{25600}{} simulation steps. 
When instantiating a policy from scratch, the BA was able to find a failure in hifi after \SI{19760}{} steps, \SI{44.1071428571429}{\percent} of the DRL baseline.
The BA was able to find a failure even faster when the policy trained in lofi was loaded, needing \SI{15230}{} steps to find a failure in hifi, \SI{33.9955357142857}{\percent} of the DRL baseline and \SI{77.0748987854251}{\percent} of the BA trained from scratch.

\begin{table}[h]
    \vspace*{0.20cm}
    \caption{The results of the time discretization case study.}
    \label{table:time_results}
    \begin{center}
        \input{Images/table_time_results}
    \end{center}
    % \vspace{-7mm}
\end{table}

\subsection{Case Study: Dynamics} \label{sec:cs_dynamics}

In this case study, the fidelity difference is in the precision of the simulator state.
The lofi simulator runs with every simulation state variable rounded to \SI{1}{} decimal point, while the hifi simulator runs with 32-bit variables.
This fidelity difference approximates when simulators may have differences in vehicle or environment dynamics.
In order to get an expert demonstration with the correct state variables, the lofi actions were run in hifi.
The results are shown in \cref{table:dynamics_results}.

The hifi DRL baseline took \SI{46800}{} simulation steps to find a failure.
The lofi solver was run for \SI{10}{} epochs, finding a failure after \SI{57200}{} simulation steps. 
When instantiating a policy from scratch, the BA was able to find a failure in hifi after \SI{13320}{} steps, \SI{28.4615384615385}{\percent} of the DRL baseline.
The BA was able to find a failure even faster when the policy trained in lofi was loaded, needing just \SI{2840}{} steps to find a failure in hifi, \SI{6.0683760683761}{\percent} of the DRL baseline and \SI{21.3213213213213}{\percent} of the BA trained from scratch.

\begin{table}[h]
    \caption{The results of the dynamics case study.}
    \label{table:dynamics_results}
    \begin{center}
        \input{Images/table_dynamics_results}
    \end{center}
    % \vspace{-7mm}
\end{table}

\subsection{Case Study: Tracker} \label{sec:cs_tracker}

In this case study, the fidelity difference is that the tracker module of the SUT perception system is turned off.
Without the alpha-beta filter, the SUT calculates its acceleration at each timestep based directly on the noisy measurement of pedestrian location and velocity at that timestep.
This fidelity difference approximates when hifi perception modules are turned off in order to achieve faster runtimes.
In order to get an expert demonstration with the correct state variables, the lofi actions were run in the hifi simulator.
The results are shown in \cref{table:tracker_results}.

The hifi DRL baseline took \SI{44800}{} simulation steps to find a failure.
The lofi solver was run for \SI{20}{} epochs, finding a failure after \SI{112000}{} simulation steps. 
When instantiating a policy from scratch, the BA was able to find a failure in hifi after \SI{18600}{} steps, \SI{41.5178571428571}{\percent} of the DRL baseline.
The BA was able to find a failure even faster when the policy trained in lofi was loaded, needing just \SI{2750}{} steps to find a failure in hifi, \SI{6.1383928571429}{\percent} of the DRL baseline and \SI{14.7849462365591}{\percent} of the BA trained from scratch.

\begin{table}[h]
    \vspace*{0.20cm}
    \caption{The results of the tracker case study.}
    \label{table:tracker_results}
    \begin{center}
        \input{Images/table_tracker_results}
    \end{center}
    % \vspace{-7mm}
\end{table}

\subsection{Case Study: Perception} \label{sec:cs_perception}

This case study is similar to the tracker case study in that it models a difference between the perception systems of lofi and hifi simulators; however, in this case study the difference is far greater.
Here, the hifi simulator of the previous case studies is now the lofi simulator.
The new hifi simulator has a perception system\footnote{Our implementation was based on that of \url{github.com/mitkina/EnvironmentPrediction}.} that uses LIDAR measurements to create a dynamic occupancy grid map (DOGMa) \cite{itkina2019dynamic,nuss2018random,hoermann2018dynamic}.
At each timestep, AST outputs the pedestrian acceleration and a single noise parameter, which is added to the distance reading of each beam that detects an object.
The SUT's perception system has \SI{30}{} beams with \SI{180}{} degree coverage and a max detection distance of \SI{100}{\meter}. 
The DOGMa particle filter uses \SI{10000}{} consistent particles, \SI{1000}{} newborn particles, a birth probability of \SI{0.02}{}, a particle persistence probability of \SI{0.99}{}, and a discount factor of \SI{0.99}{}.
Velocity and acceleration variance were initialized to \SI{12.0}{} and \SI{2.0}{}, respectively, and the process noise for position, velocity, and acceleration was \SI{0.06}{}, \SI{2.4}{}, and \SI{0.2}{}, respectively.

This case study also starts with slightly different initial conditions.
The pedestrian starting location is now \SI{2.0}{\meter} back from the edge of the road, while the vehicle starting location is only \SI{45}{\meter} from the crosswalk.
The initial velocities are the same.

The difference in noise modeling means that the action vectors lengths now differ between the lofi and hifi simulators. 
In order to get an expert demonstration, the lofi actions were run in hifi with the noise portion of the action vectors set to \SI{0}{}. 
Because the action vectors are different sizes, the solver networks have different sizes as well, so it was not possible to load the lofi policy for the BA in this case study.

The hifi DRL baseline took \SI{135000}{} simulation steps to find a failure.
The lofi solver found a failure after \SI{100000}{} simulation steps. 
The BA was able to find a failure in hifi after only \SI{6330}{} steps, a mere \SI{4.6888888888889}{\percent} of the DRL baseline.

\begin{table}[h]
    \caption{The results of the perception case study.}
    \label{table:perception_results}
    \begin{center}
        \input{Images/table_perception_results}
    \end{center}
    % \vspace{-7mm}
\end{table}

\subsection{Case Study: NVIDIA DriveSim}

As a proof-of-concept, for the final case study we implemented the new AST algorithm on a hifi simulator from industry.
Nvidia's Drivesim is a hifi simulator that combines high-accuracy dynamics with features such as perception from graphics and software-in-the-loop simulation.
An example rendering of an intersection in Drivesim is shown in \cref{fig:drivesim}.
After the AST Toolbox was connected with Drivesim, we simulated the standard crossing-pedestrian scenario (See Section IV-a) with the modified IDM as the SUT.
Here, the lofi simulator was the AST Toolbox simulator used for all previous case studies, which was trained for \SI{265450}{} steps.
Using the BA, AST was able to find a failure in \SI{4060}{} hifi steps, which took only 10 hours to run.
While the SUT was still just the modified IDM, these exciting results show that the new approach makes it possible to find failures with AST on state-of-the-art industry hifi simulators.

\subsection{Discussion}
Across every case study, a combination of running DRL in lofi and the BA in hifi was able to find failures with significantly fewer hifi steps than just running DRL in hifi directly. 
Some of the fidelity differences were quite extreme, but the BA was still able to find failures and to do so in fewer steps than were needed by just running DRL in hifi directly.
In fact, the most extreme example, the perception case study, also had the most dramatic improvement in hifi steps needed to find failure.
These results show the power of the BA in adapting to fidelity differences and make the approach of running AST in hifi significantly more computationally feasible.
Further work could explore using more traditional transfer learning and meta-learning approaches to save hifi simulation steps using lofi or previous hifi simulation training results.

The approach of loading the lofi DRL policy had interesting results as well.
In all the case studies presented here, the policy loading approach was even faster than running the BA from scratch, sometimes drastically so. 
However, throughout our work on this paper we also observed multiple cases where running the BA with a loaded policy did not result in finding failures at all, whereas running the BA from scratch was still able to find failures in those cases.
Furthermore, there are cases, for instance the perception case study in \cref{sec:cs_perception}, where loading the lofi policy is not even possible.
Based on our experiences, loading the lofi policy is a good first step: it often works, and when it works, works very well.
However, if the BA fails to find a failure with a loaded policy, then the BA should be run again from scratch, as running from scratch is a more robust failure-finding method than loading the policy from lofi. 
Future work could focus on making the BA using a loaded lofi policy more robust.
The policy has a learned standard deviation network, and one reason that the BA using a loaded lofi policy may fail sometimes is that during training in lofi the policy has already converged to small standard deviation outputs, leading to poor exploration.
Results might be improved by reinitializing the standard deviation network weights, or by finding other ways to boost exploration after a certain number of failed BA training epochs.

One final point is important to note\textemdash the goal of AST is to tractably find likely, and therefore useful, failures in a system in simulation without constraints on actor behavior that can compromise safety.
AST is not a method whose goal is to estimate the total probability of failure.
The hope of this approach is that the majority of failures in hifi are also present in lofi, with additional spurious errors, but it is certainly possible that there are some errors in hifi that have no close analog in lofi.
By biasing our search towards a likely failure found in lofi, we could actually hurt our ability to find certain hifi failures.
If our goal were to compute the total failure probability, such a bias could be a critical flaw that might lead to significantly underestimating the likelihood of failure in certain situations.
However, such a bias presents far less concern when we are instead merely looking to find a likely and instructive failure.
Indeed, the results bear this out, as the likelihoods of the failures found by the BA were not just on par with the likelihoods of the failures found by running DRL in hifi, but in fact they were actually greater across all case studies.

\section{Conclusion}

This paper presented a new algorithmic AST approach that can use data from low-fidelity interactions to reduce the number of high-fidelity simulation steps needed during training to find failures while also learning to overcome fidelity-dependent failure differences and rejecting spurious errors. 
Failures are first found in low-fidelity, and are then used as expert demonstrations for the BA to find failures in high-fidelity.
The combination of DRL in low-fidelity and the BA in high-fidelity was able to find failures faster than they were found by running DRL directly in high-fidelity across a range of case studies representing different types of fidelity differences between simulators.
The resulting speedup allows AST to be used to validate autonomous systems in high-fidelity with much faster runtimes.
%Such an improvement in AST is important in light of industry's recent commitment to using high-fidelity simulators, both in the present and the future.

\bibliographystyle{IEEEtran}
\bibliography{AST2}

\end{document}

%% file: Images/ASTStruct.tex
\begin{tikzpicture}[node distance=1.5cm,
    every node/.style={fill=white, font=\sffamily, text centered}, align=center]
  % Specification of nodes (position, etc.)
	\node (sim)             [simulator]              {Simulator $\mathcal{S}$};
    \node (solver)          [solver, above of = sim, xshift = 0.8cm]              {Solver};
    \node (reward)          [reward, above of = sim, xshift = 5cm]              {Reward\\Function};  
  % Specification of lines between nodes specified above
  % with aditional nodes for description 
  \draw[->]					(solver.west) -| node[text width=1cm, xshift = 0mm, yshift = 5mm, text centered, align=center] {Environment\\Actions} ($ (sim.north) - (15mm, 0) $);
  \draw[->] 	($(sim.east) + (0mm,2mm)$) -| node[text width=1.6cm, xshift = -17mm, yshift = 4mm] {Likelihood} ($ (reward.south) + (-2mm, 0mm) $);
  \draw[->] 	($(sim.east) + (0mm,-2mm)$) -| node[text width=1cm, xshift = -20mm, yshift = -4mm] {Event} ($ (reward.south) + (2mm, 0mm) $);
  \draw[->]		(reward.west) -- ++(0mm,0) -- node[text width=1cm, xshift = 0mm, yshift = 3mm] {Reward}(solver.east);
\end{tikzpicture}

%% file: Images/table_time_results.tex
\npdecimalsign{.}
\nprounddigits{1}
\sisetup{round-mode=places,round-precision=1}
\begin{tabular}{@{}lrrrrr@{}}
\toprule
Algorithm & \shortstack{Steps to\\Failure} & \shortstack{Final\\Reward} & \shortstack{Load Lofi\\Policy?} & \shortstack{Lofi\\Steps} & \shortstack{Percent of\\Hifi Steps}\\\midrule
BA        & \SI{19760}{}        & \SI{-794.6093}{}           & No                & \SI{25600}{}   & \SI{44.1071428571429}{\percent}           \\
BA        & \SI{15230}{}        & \SI{-745.5655}{}           & Yes               & \SI{25600}{}   & \SI{33.9955357142857}{\percent}           \\
Hifi      & \SI{44800}{}        & \SI{-819.911538649099}{}   & --                & --             & --                                         \\
\bottomrule
\end{tabular}
\npnoround

%% file: Images/table_dynamics_results.tex
\npdecimalsign{.}
\nprounddigits{1}
\sisetup{round-mode=places,round-precision=1}
\begin{tabular}{@{}lrrrrr@{}}
\toprule
Algorithm & \shortstack{Steps to\\Failure} & \shortstack{Final\\Reward} & \shortstack{Load Lofi\\Policy?} & \shortstack{Lofi\\Steps} & \shortstack{Percent of\\Hifi Steps}\\\midrule
BA        & \SI{13320}{}        & \SI{-729.747}{}             & No                & \SI{57200}{}    &    \SI{28.4615384615385}{\percent}      \\
BA        & \SI{2840}{}         & \SI{-815.84906}{}           & Yes               & \SI{57200}{}    &    \SI{6.0683760683761}{\percent}         \\
Hifi      & \SI{46800}{}        & \SI{-819.335565486123}{}    & --                & --              &    --          \\
\bottomrule
\end{tabular}
\npnoround

%% file: Images/table_tracker_results.tex
\npdecimalsign{.}
\nprounddigits{1}
\sisetup{round-mode=places,round-precision=1}
\begin{tabular}{@{}lrrrrr@{}}
\toprule
Algorithm & \shortstack{Steps to\\Failure} & \shortstack{Final\\Reward} & \shortstack{Load Lofi\\Policy?} & \shortstack{Lofi\\Steps} & \shortstack{Percent of\\Hifi Steps}\\\midrule
BA        & \SI{18600}{}        & \SI{-777.2937}{}            & No                & \SI{112000}{}         & \SI{41.5178571428571}{\percent}     \\
BA        & \SI{2750}{}         & \SI{-785.72754}{}           & Yes               & \SI{112000}{}         & \SI{6.1383928571429}{\percent}     \\
Hifi      & \SI{44800}{}        & \SI{-800.127385056536}{}    & --                & --                    & --     \\
\bottomrule
\end{tabular}
\npnoround

%% file: Images/table_perception_results.tex
\npdecimalsign{.}
\nprounddigits{1}
\sisetup{round-mode=places,round-precision=1}
\begin{tabular}{@{}lrrrrr@{}}
\toprule
Algorithm & \shortstack{Steps to\\Failure} & \shortstack{Final\\Reward} & \shortstack{Load Lofi\\Policy?} & \shortstack{Lofi\\Steps} & \shortstack{Percent of\\Hifi Steps}\\\midrule
BA        & \SI{6330}{}          & \SI{-385.7536}{}           & No                & \SI{100000}{}      & \SI{4.6888888888889}{\percent}        \\
Hifi      & \SI{135000}{}        & \SI{-511.104008557953}{}   & --                & --                 & --        \\
\bottomrule
\end{tabular}
\npnoround